\documentclass[a4paper, 10pt, twocolumn]{article}

\usepackage{amsfonts,amssymb,amsmath,amsthm}
\usepackage{graphicx}
\usepackage[footnotesize]{caption}
\usepackage{subfig}
\usepackage{epsfig,multirow}

\renewcommand{\thepage}{}

\usepackage[top=1.5cm, left=1.5cm, right=1.5cm, bottom=1.5cm]{geometry}

\title{Interpretable Deep Multimodal Image Super-Resolution}

\author{Iman Marivani, Evaggelia Tsiligianni, Bruno Cornelis, Nikos Deligiannis \\
\footnotesize \textit{Department of Electronics and Informatics, Vrije Universiteit Brussel, Pleinlaan 2, B-1050 Brussels, Belgium }\\ 
\footnotesize \textit{imec, Kapeldreef 75, B-3001 Leuven, Belgium}
}

\date{\empty} 

\renewenvironment{abstract}{\bf\small {\em\ Abstract---}}{}

\begin{document}

\maketitle

\begin{abstract} Multimodal image super-resolution (SR) is the reconstruction of a high resolution image given a low-resolution observation with the aid of another image modality.
	While existing deep multimodal models do not incorporate domain knowledge about image SR,
	we present a multimodal deep network design that
	integrates coupled sparse priors and allows the effective fusion of information
	from another modality into the reconstruction process.
	Our method is inspired by a novel iterative algorithm 
	for coupled convolutional sparse coding, resulting in an interpretable network by design.
	We apply our model to the super-resolution of near-infrared image guided by RGB images.
	Experimental results show that our model outperforms state-of-the-art methods.
\end{abstract}

\section{Introduction}
\label{sec:introduction}

Multimodal image super-resolution (SR)
refers to the reconstruction of a high-resolution (HR) image from a low-resolution (LR) observation 
using a HR image from another modality. Several studies exploit sparsity-based analytical models, e.g., coupled dictionary learning, to capture the correlation of the corresponding modalities~\cite{wang2012semi, zhuang2013supervised, deligiannis2016x, MSR}.
However, these methods suffer from
high-computational complexity of iterative algorithms,
which has been addressed by multimodal deep neural networks (DNNs)~\cite{ngiam2011multimodal, DJF, DGF}.
Multimodal deep models perform the fusion of the input signals at a shared intermediate layer, as a concatenation of the produced features of each modality~\cite{DJF, ngiam2011multimodal}.
Nonetheless, we lack a principled method to design such multimodal models 
that leverage the underlying signal structure and the correlation among modalities.

Recently, deep unfolding has been proposed for inverse problems~\cite{LISTA, hershey2014deep, ADMM-Net, raja},
i.e., implementing an iterative solution into the form of a deep network.
Inspired by analytical methods for sparse recovery, 
deep unfolding designs have been applied to several imaging problems  
to incorporate sparse priors into the solution.
Results for  image SR~\cite{Huang}, compressive imaging~\cite{zhang2018ista}, inpainting and denoising~\cite{raja}
have shown an enhanced performance by integrating domain knowledge  
into the deep networks. 
Nevertheless, these methods have been proposed for uni-modal data, therefore lacking a principled approach
to fuse information from different imaging modalities.
To the best of our knowledge, the only deep unfolding designs for multimodal image SR
have been proposed in~\cite{JMDL} and~\cite{CUNet},
that build upon existing unfolding methods for learned sparse coding~\cite{LISTA} and learned convolutional sparse coding (CSC)~\cite{raja}, respectively.  

In this paper, we present a novel multimodal deep unfolding architecture for the problem of guided image SR. Our model is inspired by a proximal algorithm for coupled convolutional sparse coding.
The corresponding iterative solution is unfolded as a convolutional neural network. The model integrates coupled convolutional sparse priors and 
allows effective fusion of the guidance modality into the reconstruction.
Unlike the existing multimodal deep learning based methods~\cite{ngiam2011multimodal, DJF, DGF},
our model is interpretable by design,
in the sense that the network operates steps of an iterative algorithm. In this paper, our method is evaluated using the task of NIR image super-resolution guided by HR RGB images. The extended results and the method description for this model can be found in~\cite{lesitaICIP}. Additionally, we have presented different network architectures for multimodal image SR applied to different guided upsampling tasks deploying several benchmark datasets in~\cite{lesitaTIP}. We have also introduced a fully connected network based on sparse coding with side information for guided image SR in~\cite{lesitaEUSIPCO}. 
\begin{figure}[t!]
	\centering
	\includegraphics[width=0.48\textwidth]{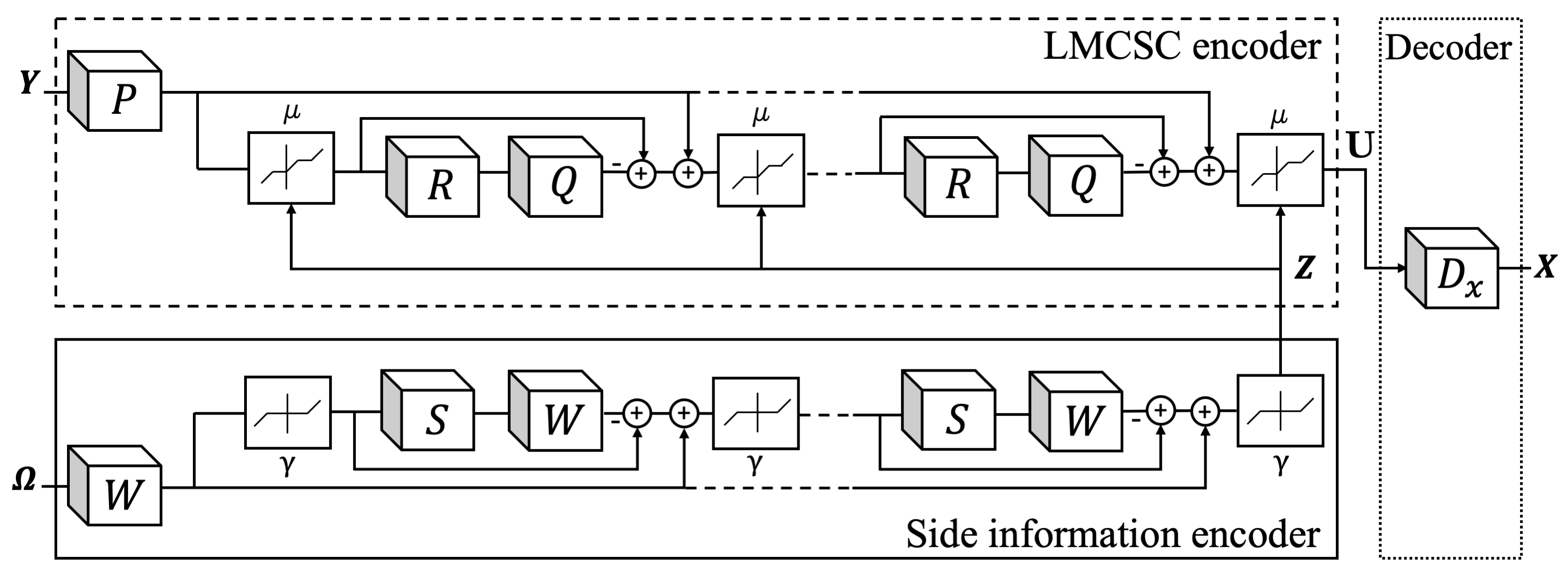}\\
	\caption{Our deep multimodal convolutional SR network. }\label{fig:MCSC}
\end{figure}

\section{The Multimodal Architecture}
\label{sec:proposed}
In this paper, we propose a multimodal architecture for guided image SR based on a novel learned multimodal convolutional sparse coding model for the fusion of the image modality.

The problem of finding coupled convolutional sparse feature maps of a target image $\boldsymbol{Y} \in \mathbb{R}^{n_1 \times n_2}$ 
with the aid of another image modality $\boldsymbol{\Omega} \in \mathbb{R}^{n_1 \times n_2}$
can be formulated as: 
\begin{equation}
\label{eq:MCSC}
\min_{\boldsymbol{\mathcal{U}}}  \frac{1}{2}\| \boldsymbol{Y} - \sum\limits_{i=1}^{k}\boldsymbol{D}_i*\boldsymbol{U}_i\|^2_2+
\lambda(\sum\limits_{i=1}^{k}\|\boldsymbol{U}_i\|_1 +\sum\limits_{i=1}^{k}\|\boldsymbol{U}_i - \boldsymbol{Z}_i\|_1),
\end{equation}
where $\boldsymbol{U}_i  \in \mathbb{R}^{n_1 \times n_2}$, $i=1,...,k$, are the sparse feature maps of  $\boldsymbol{Y}$ 
w.r.t. a convolutional dictionary $\boldsymbol{\mathcal{D}} \in \mathbb{R}^{p_1 \times p_2 \times k}$ with atoms $\boldsymbol{D}_i \in \mathbb{R}^{p_1 \times p_2}$,
and  $\boldsymbol{Z}_i  \in \mathbb{R}^{n_1 \times n_2}$, $i=1,...,k$, are the sparse feature maps of
the side information  $\boldsymbol{\Omega}$ w.r.t. a convolutional dictionary $\boldsymbol{\mathcal{B}}$ with atoms $\boldsymbol{B}_i$, 
i.e., $\boldsymbol{\Omega} = \sum_{i=1}^{k}\boldsymbol{B}_i*\boldsymbol{Z}_i$.

\begin{table*}[t!]
	\centering
	\caption{Performance comparison [in terms of PSNR (dB) and SSIM] over selected NIR test images for $\times 2$, $\times 4$ upscaling factors. }
	\label{tab:table_NIR}
	\begin{center}
		\addtolength{\tabcolsep}{-2pt}
		\scalebox{0.84}{
			\begin{tabular}{c || c |c | c| c || c| c | c |c || c |c | c |c || c |c | c |c }

				\multirow{2}{*}{Image}&\multicolumn{4}{c||}{CSCN \cite{Huang}}&\multicolumn{4}{c||}{SRFBN \cite{SRFBN}}&\multicolumn{4}{c||}{DJF \cite{DJF}}&\multicolumn{4}{c}{Proposed}\\ \cline{2-17}
				
				&\multicolumn{2}{c|}{$\times$2 }&\multicolumn{2}{c||}{$\times$4 }&\multicolumn{2}{c|}{$\times$2 }&\multicolumn{2}{c||}{$\times$4 }&\multicolumn{2}{c|}{$\times$2 }&\multicolumn{2}{c||}{$\times$4 }&\multicolumn{2}{c|}{$\times$2 }&\multicolumn{2}{c}{$\times$4 }
				\\
				\hline
				\hline
				u-0004 &   32.77& 0.9715&27.64&  0.9378&35.36& 0.9974& 29.01& 0.9787& 34.50& 0.9964& 31.02& 0.9784&  \textbf{37.36}&  \textbf{0.9977}& \textbf{33.75} & \textbf{0.9869}\\
				u-0006 &   39.47& 0.9715&32.60& 0.9361&41.08& 0.9970& 33.73& 0.9702& 41.52& 0.9975& 36.04& 0.9894&  \textbf{43.60}&  \textbf{0.9982}& \textbf{38.74} & \textbf{0.9912}\\
				u-0017 &   36.76& 0.9574&32.60& 0.9361&38.19& 0.9950& 32.91& 0.9725& 38.65& 0.9961& 34.18& 0.9815&  \textbf{40.87}&  \textbf{0.9967}& \textbf{36.16} & \textbf{0.9828}\\
				o-0018 &   33.98& 0.9659&27.28& 0.9250&36.47& 0.9971& 28.88& 0.9740& 34.78& 0.9960& 30.72& 0.9888&  \textbf{39.21}& \textbf{0.9982}& \textbf{34.17} & \textbf{0.9902}\\
				u-0020 &   35.54& 0.9658&30.04& 0.9487&37.50& 0.9969& 31.44& 0.9807& 37.35& 0.9973& 33.60& 0.9915& \textbf{40.98}&  \textbf{0.9980}& \textbf{36.95} & \textbf{0.9900}\\
				u-0026 &   32.94& 0.9339&27.91& 0.8724&31.00& 0.9782& 29.10& 0.9702& 33.15& 0.9939& 29.21& 0.9397&  \textbf{35.60}&  \textbf{0.9963}& \textbf{31.03} & \textbf{0.9784}\\
				o-0030 &   33.34& 0.9465&27.72& 0.8378&35.57& 0.9944& 29.45& 0.9583& 35.67& 0.9944& 31.27& 0.9345& \textbf{38.29}&  \textbf{0.9961}& \textbf{33.56} & \textbf{0.9780}\\
				u-0050 &   33.31& 0.9693&28.20& 0.9101&37.06& 0.9966& 29.89& 0.9762& 32.60& 0.9928& 28.58& 0.9616& \textbf{34.11}&  \textbf{0.9948}& \textbf{30.04} & \textbf{0.9772}\\
				\hline
				\hline
				Average &  34.76& 0.9602&29.14& 0.9111&36.53&  0.9941& 30.55& 0.9726& 36.03& 0.9955& 31.83& 0.9707&  \textbf{38.74}&  \textbf{0.9970}& \textbf{34.28} & \textbf{0.9843}\\
				
		\end{tabular}}
	\end{center}
\end{table*}
Considering the linear properties of convolution, it is easy to write \eqref{eq:MCSC} in the form of $\ell_1$-$\ell_1$ minimization~\cite{nikosIT}:
\begin{equation}
\label{l1l1}
\min_{\boldsymbol{x}}  \frac{1}{2}\| \boldsymbol{y} - \boldsymbol{A}\boldsymbol{x}\|^2_2+\lambda(\|\boldsymbol{x}\|_1 +\|\boldsymbol{x}-\boldsymbol{s}\|_1),
\end{equation}
Where the dictionary $\boldsymbol{A} \in \mathbb{R}^{(n_1-p_1+1)(n_2-p_2+1) \times kn_1n_2} $ is defined
as a concatenation of Toeplitz matrices that unroll $\boldsymbol{D}_i$'s;
$\boldsymbol{y} \in \mathbb{R}^{n_1n_2}$, $\boldsymbol{x} \in \mathbb{R}^{kn_1n_2}$ and $\boldsymbol{s} \in \mathbb{R}^{kn_1n_2}$ are vectorized forms of the LR image $\boldsymbol{Y}$ and the sparse feature maps 
of the target and the guidance images, respectively.
This reformulation allows us to utilize the proximal algorithm given by \cite{lesitaSPL} to solve~\eqref{l1l1} as:
\begin{equation}
\label{eq:lesita}
\boldsymbol{x}^{t+1} = \xi_{\mu}\big(\boldsymbol{x}^t- \frac{1}{L}\boldsymbol{A}^T\boldsymbol{A}\boldsymbol{x}^t + \frac{1}{L}\boldsymbol{A}^T\boldsymbol{y};\boldsymbol{s}\big),
\end{equation}

However, since the coupled convolutional sparse coding deals with an entire image, using \eqref{eq:lesita} to solve (\ref{eq:MCSC}) is not practical.
Instead, we describe \eqref{eq:lesita} in a convolutional form 
based on the following observations:
Considering the correspondence between the special structure of~$\boldsymbol{A}$ 
and the convolution operator,
multiplying $\boldsymbol{A}$ with the sparse codes $\boldsymbol{x}$ in \eqref{eq:lesita}
is equivalent to the calculation of the sum of convolutions in (\ref{eq:MCSC}). 
Since the transpose of a Toeplitz matrix is also Toeplitz, 
the terms including $\boldsymbol{A}^T$ in~\eqref{eq:lesita} can also be seen as convolutions. 
By replacing the matrix-vector multiplications with convolutions, (\ref{eq:lesita}) be described as
\begin{equation}
\label{eq:lesita55}
\boldsymbol{\mathcal{U}}^{t+1} = \xi_{\mu}(\boldsymbol{\mathcal{U}}^t - \boldsymbol{\mathcal{\tilde{D}}}*\boldsymbol{\mathcal{D}}*\boldsymbol{\mathcal{U}}^t+\boldsymbol{\mathcal{\tilde{D}}}*\boldsymbol{Y};\boldsymbol{\mathcal{Z}}),
\end{equation}
where $\boldsymbol{\mathcal{\tilde{D}}}$ denotes a convolutional dictionary corresponding to $\boldsymbol{A}^T$. We translate the iterative solution given by \eqref{eq:lesita55} into a deep convolutional neural network (CNN);
each stage of the network calculates the sparse feature maps based on  
\begin{equation}
\label{eq:lesita3}
\boldsymbol{\mathcal{U}}^{t+1} = \xi_{\mu}(\boldsymbol{\mathcal{U}}^t - \boldsymbol{\mathcal{Q}}*\boldsymbol{\mathcal{R}}*\boldsymbol{\mathcal{U}}^t+\boldsymbol{\mathcal{P}}*\boldsymbol{Y};\boldsymbol{\mathcal{Z}}),
\end{equation}
where the convolutional filters $\boldsymbol{\mathcal{Q}}$, $\boldsymbol{\mathcal{R}}$, $\boldsymbol{\mathcal{P}}$ are learnable and $\mu$ is a learnable proximal parameter. We refer to this network as Learned Multimodal Convolutional Sparse Coding (LMCSC).

The proposed multimodal architecture for guided image SR, which is shown in Fig.~\ref{fig:MCSC}, is built using three subnetworks:
(\textit{i}) an LMCSC encoder that calculates latent convolutional representations of the LR from the target image modality with the aid of side information.
map $\boldsymbol{\mathcal{Z}}  \in \mathbb{R}^{n_1 \times n_2 \times k}$ as side  information
(\textit{ii}) a convolutional LISTA encoder \cite{raja} that provides the latent features of the guidance image (which serve the side information for the LMCSC branch),
and (\textit{iii}) a decoder in the form of a convolutional dictionary that reconstruct the HR image.

The entire network is trained end-to-end by minimizing the reconstruction error between the reconstructed and the ground truth HR\ image from the target modality. 

\begin{figure}
	\centering
	\subfloat[CSCN\cite{Huang}]{\includegraphics[scale=0.14]{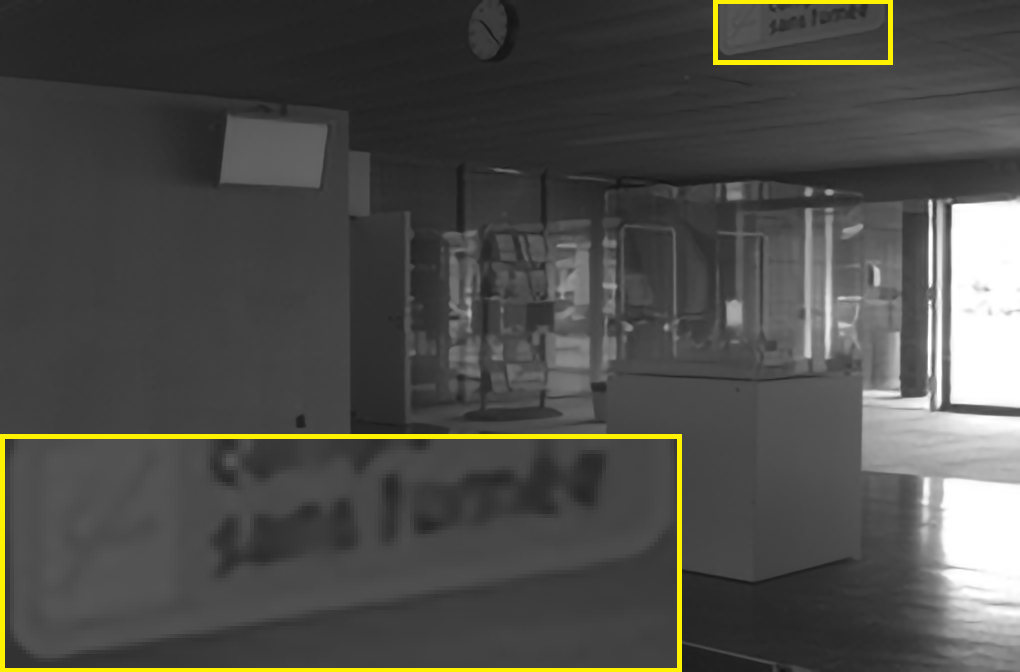}}
	\hspace{0.05cm}
	\subfloat[SRFBN~\cite{SRFBN}]{\includegraphics[scale=0.14]{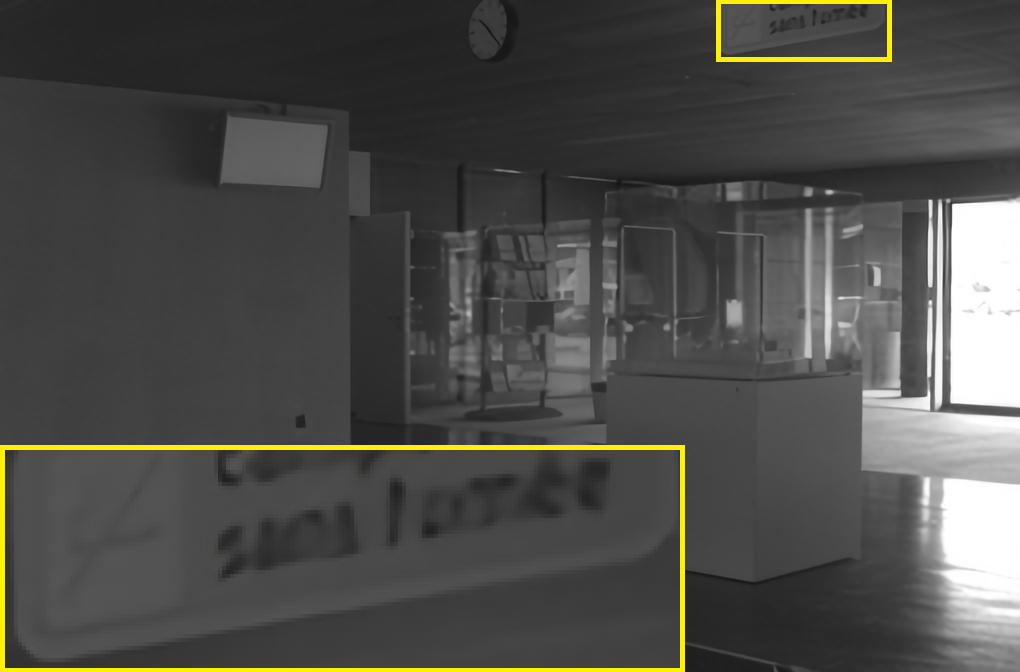}}
	\hspace{0.05cm}
	\subfloat[DJF~\cite{DJF}]{\includegraphics[scale=0.14]{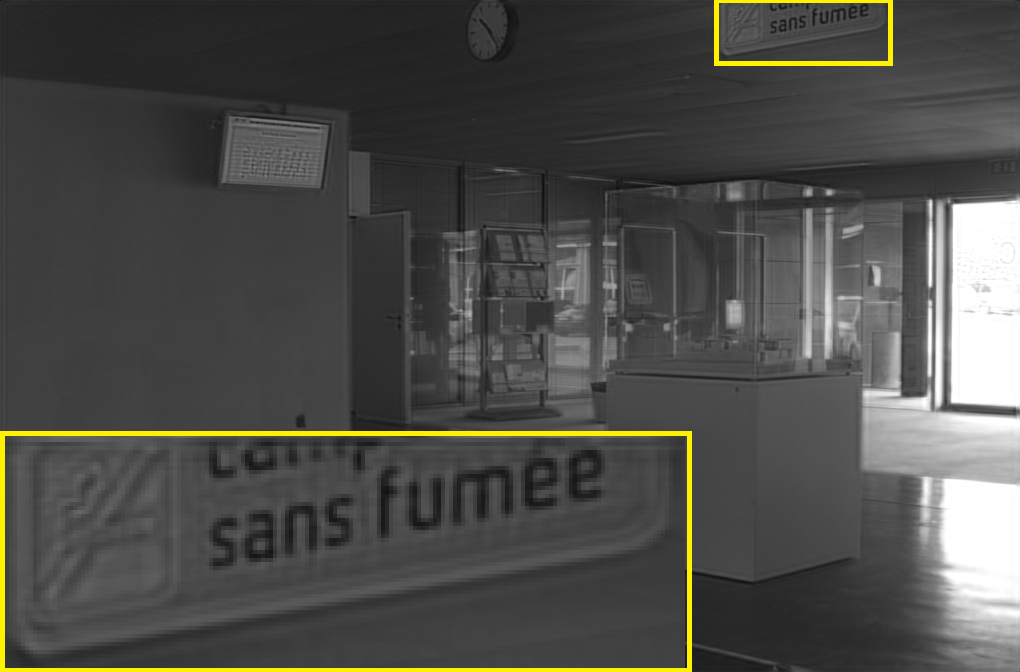}}
	\hspace{0.05cm}
	\subfloat[ Proposed]{\includegraphics[scale=0.14]{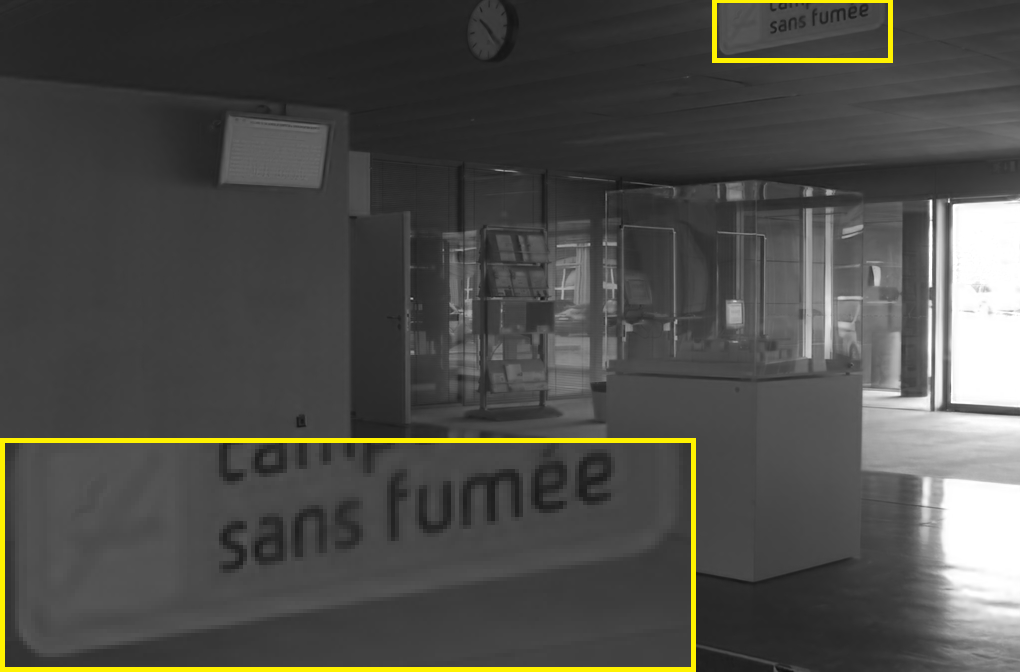}}
	\hspace{0.05cm}
	
	\caption{$\times$4 scale super-resolution of the test image \textquotedblleft{u-0013}\textquotedblright.}   
	\label{fig:visualEx}
\end{figure}

\section{Experiments}
\label{sec:experiments}
This section presents the implementation details for the proposed model 
and its performance evaluation. We employ the EPFL RGB-NIR dataset\footnote{https://ivrl.epfl.ch/supplementary\_material/cvpr11/}
for the experiments on guided NIR image super-resolution. 
We create a training set consisting of $40,000$ image patches 
of size $64\times64$ extracted from the original images. 
We produce the LR version of the ground truth images from the target modality 
by performing blurring and downscaling operations. 
The guidance modality in our network only includes the luminance channel of the corresponding RGB image. 
We reserve seven pairs for testing.
We apply SR at different scales,
and train the network separately for every scale.

The convolutional layers contain $85$ kernels of size $5\times5$. 
Zero padding is applied to the input of each convolutional layer to preserve the same spatial size throughout the model. 
The convolutional kernels are initialized using a Gaussian distribution with a standard deviation $0.01$. 
The proximal parameters $\mu$ and $\gamma$ are initialized to $0.2$. 
We train the network using the Adam optimizer with a mini-batch size $32$.

We compare the performance of our network against several SR models including, CSCN~\cite{Huang}, SRFBN~\cite{SRFBN} and the deep joint image filtering (DJF)~\cite{DJF} models. The experimental results for  different upscaling factors, are summarized in Table~\ref{tab:table_NIR}. It is clear that the proposed model notably outperforms the state-of-the-art, bringing average peak signal-to-noise-ratio (PSNR) improvements of up to 2.7dB compared to the best performing competing method. These improvements are also corroborated by visual examples depicted in Fig.~\ref{fig:visualEx}.

\section{Conclusion}
\label{sec:conclusion}
In this paper, we presented our multimodal deep architecture for guided image super-resolution. The architecture is designed based on a novel deep unfolding model that learns the coupled convolutional sparse representations. We deploy the multimodal network to super-resolve NIR\ images using high-resolution RGB\ images as side information. The experimental results demonstrate the high reconstruction accuracy of the proposed model outperforming several state-of-the-art methods.

\bibliographystyle{IEEEbib}
{\footnotesize \bibliography{IEEEabrv,refs}}

\end{document}